\documentclass[letterpaper, 10 pt, conference]{ieeeconf}  

\IEEEoverridecommandlockouts                              

\usepackage{ral}
\usepackage{subcaption}
\usepackage{cite}
\usepackage{hyperref}

\definecolor{somegray}{rgb}{0.5, 0.5, 0.5}
\newcommand{\darkgrayed}[1]{\textcolor{somegray}{#1}}
\makeatletter
\newcommand*\titleheader[1]{\gdef\@titleheader{#1}}
\AtBeginDocument{%
  \let\st@red@title\@title
  \def\@title{%
    \vskip-4em
    \bgroup\normalfont\large\centering\@titleheader\par\egroup
    \vskip1.5em\st@red@title}
}

\titleheader{\darkgrayed{This paper has been accepted for publication at the
\\ IEEE International Conference on Robotics and Automation (ICRA), Atlanta, 2025.}}

\title{\LARGE \bf
Learning Quadrotor Control From Visual Features
\\Using Differentiable Simulation
}

\author{
    Johannes Heeg, Yunlong Song, Davide Scaramuzza
    \thanks{
    The authors are with the Robotics and Perception Group, Department of Informatics, University of Zurich, Switzerland (\protect\url{http://rpg.ifi.uzh.ch}). This work was supported by the European Union’s Horizon Europe Research and Innovation Programme under grant agreement No. 101120732 (AUTOASSESS) and the European Research Council (ERC) under grant agreement No. 864042 (AGILEFLIGHT).}
}

\begin{document}
\maketitle
\thispagestyle{empty}
\pagestyle{empty}

\begin{abstract}

The sample inefficiency of reinforcement learning~(RL) remains a significant challenge in robotics.
RL requires large-scale simulation and can still cause long training times, slowing research and innovation. 
This issue is particularly pronounced
in vision-based control tasks where reliable state estimates are not accessible
Differentiable simulation offers an alternative by enabling gradient back-propagation through the dynamics model, providing low-variance analytical policy gradients and, hence, higher sample efficiency.
However, its usage for real-world robotic tasks has yet been limited.  
This work demonstrates the great potential of differentiable simulation for learning quadrotor control. 
We show that training in differentiable simulation significantly outperforms model-free RL in terms of both sample efficiency and training time, allowing a policy to learn to recover a quadrotor in seconds when providing vehicle states and in minutes when relying solely on visual features.  
The key to our success is two-fold. 
First, the use of a simple surrogate model for gradient computation greatly accelerates training without sacrificing control performance. 
Second, combining state representation learning with policy learning enhances convergence speed in tasks where only visual features are observable. 
These findings highlight the potential of differentiable simulation for real-world robotics and offer a compelling alternative to conventional RL approaches.
\end{abstract}

\vspace{0.2cm} 
\noindent
\textbf{Video:} \href{https://youtu.be/LdgvGCLB9do}{https://youtu.be/LdgvGCLB9do} \\
\textbf{Code:} \href{https://github.com/uzh-rpg/rpg_flightning}{https://github.com/uzh-rpg/rpg\_flightning}
\section{Introduction}

Reinforcement Learning (RL) has become a cornerstone in robotics research for designing powerful controllers.
By employing large-scale simulation, neural network controllers trained with RL enabled impressive performance in many robotic tasks~\cite{tang2024rlrobotics} like quadruped locomotion~\cite{hwangbo2019learning}, robot soccer~\cite{haarnoja2024}, or quadrotor control~\cite{hwangbo2017control, song2021autonomousdrone}.
The learning paradigm allows for the end-to-end design of control policies that take various input modalities and optimize task-level objectives~\cite{song2023reaching}. However, the flexibility comes at the cost of low sample efficiency, which slows down the development of new controllers.
Successful implementations require massively parallelized simulators and large compute resources. Yet, training times can go up to hours or even days.
This issue makes iterating on the control architecture, environment settings, and hyperparameter tuning tedious and delays new results and progress.

The sample efficiency problem becomes even more pronounced in vision-based control tasks~\cite{geles2024demonstrating}.
Not only is the simulation more expensive due to the additional requirement of simulating the visual observations, a direct mapping from visual input signals to control commands is more difficult to learn.
First, the dimensionality of the inputs increases, requiring more samples to optimize the policy.
Second, the neural network must simultaneously build an implicit state representation and synthesize the control law.

Differentiable simulation offers an approach to improve the sample efficiency in robot learning.
By incorporating the derivatives of the dynamics, policy gradients can be computed analytically.
This method reduces the variance in the policy parameter updates, which potentially decreases the number of samples needed to learn a new control task.

\begin{figure}[tp]
    \centering
    
    \begin{subfigure}[b]{\linewidth}
        \centering
        \includegraphics[width=\linewidth]{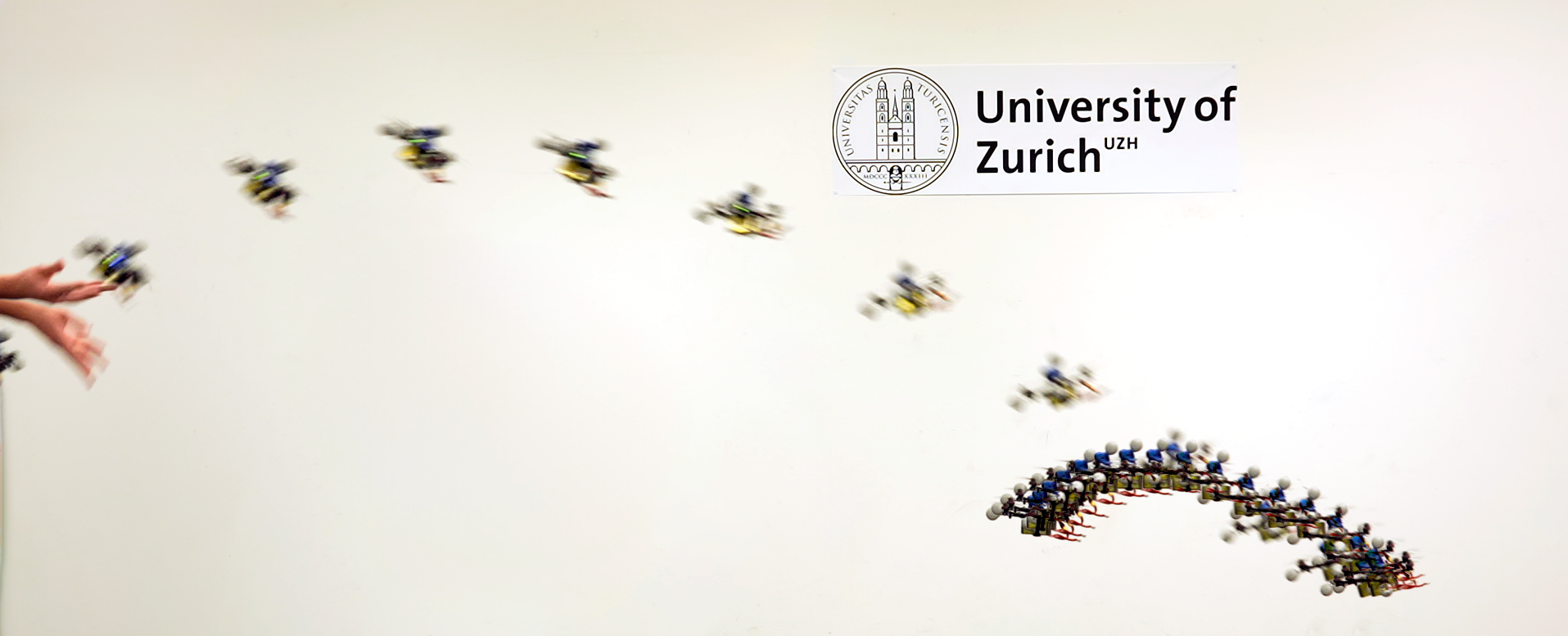}
    \end{subfigure}

    \begin{subfigure}[b]{0.49\linewidth}
        \centering
        \includegraphics[width=\linewidth]{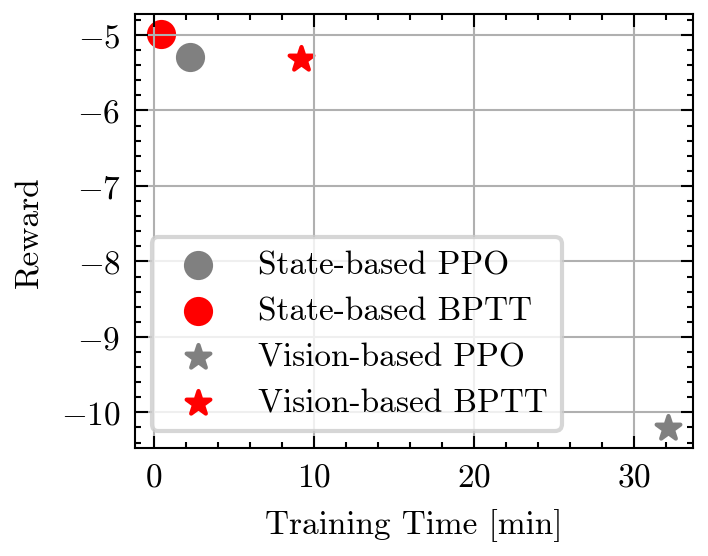}
    \end{subfigure}
    \hfill
    \begin{subfigure}[b]{0.49\linewidth}
        \centering
        \includegraphics[width=0.8\linewidth]{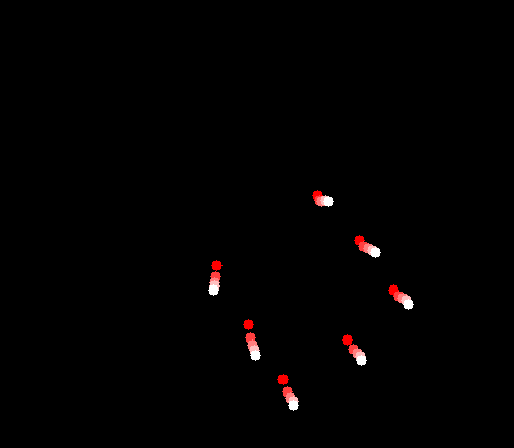}
    \end{subfigure}
    
    \caption{\textbf{Learning quadrotor control from visual features.} 
    \emph{Top}: Stabilizing a quadrotor from a hand throw. 
    \emph{Bottom Left:} Backpropagation through time (BPTT) via differentiable simulation outperforms PPO, enabling state-based control in seconds and vision-based control in minutes. 
    \emph{Bottom Right:} A visualization of observed visual features using hardware-in-the-loop simulation.}
    \label{fig:fig1}
\end{figure}

\begin{figure*}[t]
    \centering
    \includegraphics[width=0.9\linewidth]{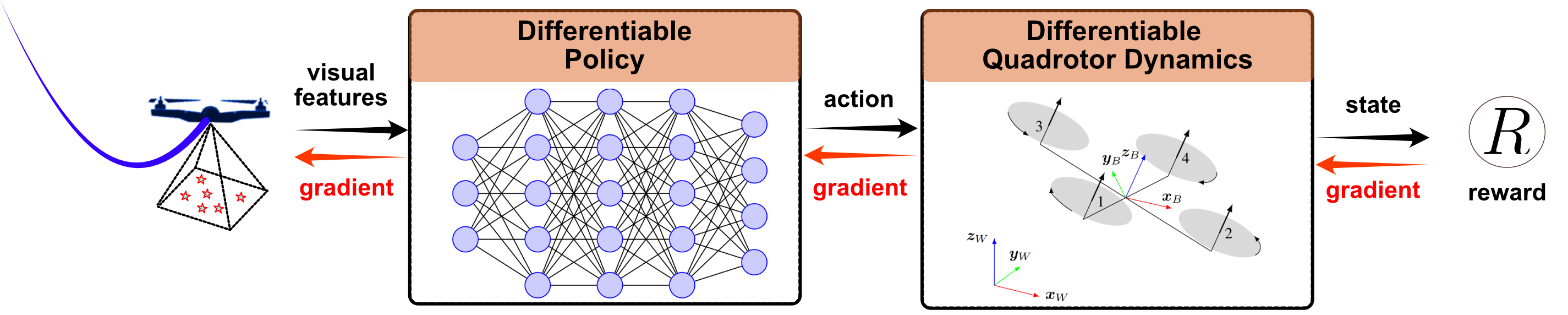}
    \caption{Overview of policy training using differentiable simulation. A neural network policy takes actions based on visual feature observations. The simulation state is then updated using differentiable quadrotor dynamics. Based on the next state, new observations are computed with a differentiable camera model. For each state and action pair, the actor receives a reward. The differentiability of the whole pipeline allows back-propagating the gradients from the rewards to the policy parameters.}
    \label{fig:overview}
\end{figure*}

We investigate using differentiable simulation to learn vision-based quadrotor control.
In our case study, the objective is to stabilize a quadrotor from a manual throw using only visual features extracted from camera observations (see \autoref{fig:fig1}). The task is challenging for several reasons.
First, quadrotors are inherently underactuated and unstable systems; without continuous and precise feedback control, they quickly lose stability and crash.
Second, state information is only indirectly available through pixel coordinates in the image plane, making the environment partially observable.
Finally, the control policy has to handle a wide initial state distribution to recover from a manual throw.

In our study, we begin with a classical state-based control task and then transition to vision-based control directly from visual features (i.e., without state estimation).
We compare learning in differentiable simulation with PPO (Proximal Policy Optimization)~\cite{schulman2017ppo}, a popular model-free RL algorithm.
Remarkably, after just 23 seconds of training, our state-based control policy trained in differentiable simulation successfully learns to stabilize the quadrotor from random initial configurations. 
In comparison, PPO requires over five times more samples and training time to achieve similar results.
For feature-based control, our network trained in differentiable simulation learns to stabilize the quadrotor based on observations from visual features only. 
In contrast, PPO, even after extensive training, converges to a sub-optimal reward that is significantly lower than that of differentiable simulation, highlighting its limitations in vision-based tasks.

\textbf{Contribution:} 
The key contribution of our work is leveraging differentiable simulation for quadrotor control, focusing on controlling the vehicle from visual features only. 
We demonstrate that, first, differentiable simulation offers a significant advantage over model-free RL for both state-based and vision-based control tasks.
Second, incorporating a simplified surrogate model in backward pass speeds up gradient computation without compromising on sample efficiency and performance. 
Third, pretraining the control policy on an auxiliary representation task accelerates and stabilizes the overall training process.
\section{Related Work}

Model-free RL is popular method to learn quadrotor control policies, including stabilization~\cite{hwangbo2017control, eschmann2024learning} and high-speed drone racing~\cite{song2023reaching, kaufmann2023champion}.
For vision-based control, it is common to split the design process into state estimator and controller design.
In~\cite{kaufmann2023champion}, the quadrotor state is estimated using a Kalman filter that receives measurements from a visual-inertial odometry module and additional visual features.
The state estimate is then fed into a state-based policy that was trained using RL.
Splitting state estimation and control gives rise to a modular structure.
However, the composition can lead to new failure modes caused by compounding errors.
Alternatively, an end-to-end learning approach is studied in~\cite{geles2024demonstrating} for drone racing. The authors demonstrate robust flight performance solely relying on visual inputs.
Their experiments show that learning from visual inputs requires much more samples and longer training times of around 24 hours.

Most studies on differentiable simulation for policy optimization in robotics have been limited to simulated experiments~\cite{freeman2021brax, xu2021accelerated, howelllecleach2022}, with only a few successes in real-world applications, such as learning quadruped locomotion~\cite{song2024learning} and vision-based drone swarm navigation~\cite{zhang2024back}.
In particular,~\cite{zhang2024back} demonstrates how a simple point-mass model can be used to learn vision-based navigation policies for agile flight in complex real-world environments. 
Their approach requires state estimation and a low-level controller that can stabilize the quadrotor and track the acceleration commands from the policy.
Hence, there remains a research gap in exploring low-level quadrotor control that fully incorporates the complete vehicle dynamics, which is important for leveraging the full potential of a quadrotor for agile maneuvers, such as acrobatic flight or time-optimal flight.

Using a surrogate model for policy gradient computations was first studied in~\cite{abbeel2006}.
The authors propose using an approximate simulation model to compute gradients along real-world trajectories.
They prove that the resulting policy parameters lie close to a local optimum with respect to the true but unknown model.
Another work~\cite{kolter2009}~even considers the case in which the surrogate model only predicts the sign of the derivatives.
Recently, \cite{song2024learning}~proposed a surrogate model for gradient computations in the context of quadruped locomotion.
\section{Method}

The key benefit of differentiable simulators is the possibility of back-propagating gradients from objective functions through the dynamics to optimization parameters.
This feature can be used to train control policies with analytically derived gradients.
Our fully differentiable simulation pipeline for vision-based control is summarized in \autoref{fig:overview}.
In the following, we introduce how differentiable simulation can be used for policy optimization and describe the pipeline in more detail.

\subsection{Policy Optimization via Differentiable Simulation}

The quadrotor is modeled as a discrete-time dynamical system, characterized by continuous state 
and control input spaces, denoted as \( \mathcal{X}  \) and \( \mathcal{U} \), 
respectively. At each time step \( t \), the system state is \( x_t \in 
\mathcal{X} \), and the corresponding control input is \( u_t \in \mathcal{U} 
\).
An observation \( o_t \in \mathcal{O} \) is generated at each time step based on 
the current state \( x_t \) through an observation model \( h: \mathcal{X} \rightarrow 
\mathcal{O} \), such that \( o_t = h(x_t) \).
The system dynamics are governed by the function \( f: \mathcal{X} \times 
\mathcal{U} \rightarrow \mathcal{X} \), which describes the time-discretized 
evolution of the system \( x_{t+1} = f(x_t, u_t) \). 
At each time step \( t \), the robot receives a reward signal \( r_t = r(x_t, u_t)\). 
The control policy is a deterministic differentiable function, a neural network,~\( u_t = \pi_{\theta} (o_t)\). The policy takes the observation \( o_t \) as the input and outputs the control input \( u_t \).

The objective function $R(\theta)$, the cumulative reward given the policy parameters, is to be maximized to find the optimal policy parameters 
\(\theta^{\ast}\) by means of gradient ascent
\begin{align*}
\theta^{\ast} &= \arg \max_{\theta} R(\theta) \\
R(\theta)&=  
  \sum_{t=0}^{N-1} r(x_t, u_t) 
  = 
  \sum_{t=0}^{N-1} r(x_t, \pi_{\theta}(o_t)) \\
  &= \sum_{t=0}^{N-1} r(x_t, \pi_{\theta}(h(x_t)))
  \\
  \theta_{k+1} & \leftarrow \theta_k + \alpha \nabla_{\theta} R(\theta_k),
\end{align*}
where $\alpha$ is the learning rate. 

\textbf{Back-propagation Through Time (BPTT):}
By taking advantage of the structure of $R(\theta)$ and applying the chain rule along the temporal axis, the policy gradient is obtained through BPTT and reads
\begin{align*}
    \nabla_{\theta} R(\theta) &= \\
     \frac{1}{N}& \sum_{t=0}^{N-1} \left( \sum_{i=1}^{t} \frac{\partial r_t 
    }{\partial x_t} \prod_{j=i}^{t} \left( \frac{d x_{j}}{d 
  x_{j-1}} \right) \frac{\partial x_i}{\partial \theta} + \frac{ \partial 
r_t}{\partial u_t} \frac{ \partial u_t}{\partial \theta} \right),
\label{eq: first_order_gradient}
\end{align*}
where the matrix of derivatives \( d x_j / d x_{j-1}\) is 
the Jacobian of the close-loop dynamical system~$f(x, \pi_\theta(h(x)))$.
For a derivation of the BPTT gradient, the reader is referred to~\cite{metz2021gradients}.
%

\subsection{Quadrotor Dynamics}

Let \(\bm{p}\), \(\bm{R}\), and \(\bm{v}\) represent the position, orientation matrix, and linear velocity of the quadrotor, respectively, expressed in the world frame. 
Let \(\bm{\omega}\) denote the angular velocity of the quadrotor expressed in the body frame, referred to as body rates. 
Additionally, let \(c\) represent the mass-normalized collective thrust produced by all motors, and let \(\bm{c} = \begin{bmatrix} 0 & 0 & c\end{bmatrix}^\intercal\) denote the collective thrust vector and \(\bm{g}\) the gravity vector.
The simple quadrotor dynamics read
\begin{equation}
	\begin{aligned}
		\dot{\bm{x}} = \frac{d}{dt}\begin{bmatrix}
   		  \bm{p} \\
			\operatorname{vec}(\bm{R}) \\
			\bm{v} \\
		\end{bmatrix} = \begin{bmatrix}
		\bm{v}\\
		\operatorname{vec}(\mathbf R [\bm \omega]_\times)\\
		\bm{R} \bm{c} + \bm{g}\\

	\end{bmatrix},
	\end{aligned}
 \label{eq:dynamics}
\end{equation}where $[\cdot]_\times$ is the skew symmetric matrix operator and $\operatorname{vec}(\cdot)$ indicates vectorization.
For quadrotor control, the body rates \(\bm{\omega}\) and the collective thrust \(c\) are the inputs to the system.

\subsection{JAX-based Differentiable Simulator}
\label{sec:jax}

To enable back-propagating gradients through simulated rollouts and exploit GPU-accelerated computing, the differentiable simulation framework is entirely written using JAX~\cite{jax2018github}. During runtime, the code is just-in-time compiled and lowered to GPU, resulting in fast execution times of up to million of steps per second (\autoref{tab:sim-speed}). Besides high learning speeds, JAX supports automatic differentiation which can be used to compute analytical policy gradients.

Inspired by the quadrotor simulator presented in~\cite{song2020flightmare}, our differentiable simulator models air-drag, the low-level control architecture, the motor speeds, and the transmission delays. This fidelity allows the transfer of policies trained in simulation to the real world zero-shot.

\subsection{Fast Surrogate Gradients}

Even though our simulator is fully differentiable, automatic differentiation through a full low-level control stack is not desirable. The high simulation frequency of the dynamics model, 1000 Hz, and the different sub-models involved result in an expansive computational graph for gradient computation, leading to slower execution and larger memory demand.

To circumvent this problem while keeping the high fidelity of our simulator, we use the simple dynamics model $\hat f(x, u)$ (see equation \eqref{eq:dynamics}), to compute the gradients. While during the forward path, the full model
$$ x_{t+1} = f(x_t, u_t) $$ is used, during backpropagation, we set
$$ \frac{\partial x_{t+1}}{\partial x_t} := \left.\frac{\partial \hat f}{\partial x}\right|_{(x_t, u_t)}, \quad \frac{\partial x_{t+1}}{\partial u_t} := \left.\frac{\partial \hat f}{\partial u}\right|_{(x_t, u_t)}. $$
This technique combines the realistic forward model with a computationally cheap backward model and keeps the overhead of BPTT low (\autoref{tab:sim-speed}).

\begin{table}
\centering
\setlength\tabcolsep{4pt}
\small
\renewcommand*{\arraystretch}{1.2}
\fontsize{7pt}{8pt}\selectfont
\begin{tabular}{r|rrr}
\toprule
\multicolumn{1}{c}{\multirow{2}{*}{\textbf{Parallelized Environments}}} & \multicolumn{3}{c}{\textbf{Steps per Second}}                                                                                         \\ 
\multicolumn{1}{c}{}                                & \multicolumn{1}{c}{Rollout} & \multicolumn{1}{c}{BPTT Simple} & \multicolumn{1}{c}{BPTT Full} \\ \midrule
1                                          & 1,539                                & 835                                      & 234                                    \\
10                                         & 12,972                               & 7,318                                    & 2,221                                  \\
100                                        & 98,919                               & 59,767                                   & 19,432                                 \\
1,000                                      & 875,082                              & 489,927                                  & 165,425                                \\
10,000                                     & 4,907,277                            & 2,132,523                                & \multicolumn{1}{c}{$-$}                                       \\
100,000                                    & 7,227,766                            &  \multicolumn{1}{c}{$-$}                   &  \multicolumn{1}{c}{$-$}                                      \\ \bottomrule
\end{tabular}
\caption{Simulation speeds with and without gradient computation (BPTT) on an Nvidia Titan RTX (24 GB VRAM). Using a simple model for the backward path (BPTT simple) keeps computation and memory costs low. For empty entries, the computational graph required more memory than accessible.}
\label{tab:sim-speed}
\end{table}

\subsection{Pretraining on State Representation Learning}

Learning from visual features comes with additional challenges for policy optimization using BPTT. First, the function mapping from features to commands is more difficult for a neural network to approximate. Second, the observation model causes indirect gradient flow from the actions back to the states.
Both challenges presumably decrease the sample efficiency.
We propose to pretrain the neural network parameters on the state representation task, which is known to improve the convergence of reinforcement learning~\cite{debruin2018representation}.

The pretraining procedure comprises three steps.
First, a randomly initialized policy network $\pi_\theta$ is used to collect a dataset $\mathcal D$ of pairs of simulation states and observations $(x_t, o_t)$. Then, a neural network is trained to predict the underlying simulation state (the quadrotor state) from the visual observations.
I.e., the neural network $\psi_\theta$ is trained to minimize the mean squared error
$$\frac{1}{N} \sum_{(x_i, o_i) \in \mathcal{D}}\|x_i - \psi_\theta(o_i)\|^2.$$

Lastly, the resulting weights $\theta$ are copied to the policy network $\pi_\theta$.
Note that the last layer is changed do to the different output modalities.

\subsection{Training Details}

\subsubsection{Policy}
We parameterize the policy as a multi-layer perceptron with two hidden layers.
The size of the hidden layers is 512 for state-based and 1024 for vision-based control.

\subsubsection{Action Space}
The policy network provides actions at 50 Hz to the on-board flight controller. The actions are 4-dimensional and consist of desired mass-normalized collective thrust and body rates 
$u = [c, \omega_x, \omega_y, \omega_z]^\top.$
This control modality keeps full control authority to the policy network and does not need state estimation on the flight controller. On the other hand, it requires the policy to not only steer the quadrotor but rather stabilize it at all times which is not-trivial given the non-linear, unstable dynamics~\eqref{eq:dynamics}.

\begin{figure}[t]
    \centering
    \includegraphics[width=0.5\linewidth]{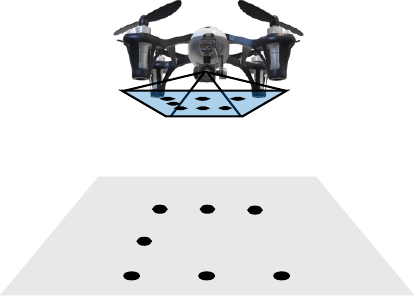}
    \caption{Vision-based control. The actor observers only a history of pixel coordinates of seven points on the ground and the last three actions taken.}
    \label{fig:feaeture-obs}
\end{figure}

\subsubsection{Observation Space}
\label{sec:obs}
In our first experiment, we assume the quadrotor state to be known resulting in an observation
$o = [\mathbf{p}^\top, \operatorname{vec}(\mathbf{R})^\top, \mathbf{v}]^\top$.
In our second experiment, learning to fly from visual features, the policy observes the pixel coordinates of 7 features on the ground as indicated in \autoref{fig:feaeture-obs}. Since it is impossible to estimate velocities and accelerations from only one frame, we include the pixel coordinates from the past 5 frames and the last 3 actions taken by the policy.
The pixel coordinates are obtained through a realistic camera model \cite{usenko2018doublesqhere}. We implemented the camera model in a differentiable fashion, allowing gradients to be propagated through the observations.

\subsubsection{Reward Function}
Stabilizing the quadrotor means regulating its state toward a hovering condition at a desired position $\bm{p}_\mathrm{des}$.
The desired behavior is encoded into the reward function. At time $t$

$$r_t = r_t^{\mathrm{pos}} + r_t^{\mathrm{vel}} + r_t^{\mathrm{act}}.$$
where the individual terms are
\begin{align*}
    r_t^{\mathrm{pos}} &= -0.2 \cdot L_\mathrm{H}(5 \cdot (\bm{p}_t - \bm{p}_\mathrm{des}))\\
    r_t^{\mathrm{vel}} &= -0.1\cdot L_\mathrm{H}(\bm{v}_t) - 0.1 \cdot L_\mathrm{H}(\bm{\omega}_t)\\
    r_t^{\mathrm{act}} &= -0.5\cdot L_\mathrm{H}(\bm{u}_t - \bm{u}_\mathrm{hover}) 
- 0.01\cdot L_\mathrm{H}(\bm{u}_t - \bm{u}_{t-1}).
\end{align*}
Here, $L_\mathrm{H}$ is the Huber loss and $\bm{u}_\mathrm{hover} = [9.81, 0,0,0]^\top$, the action required to withstand the gravity.
We use the Huber loss to limit the magnitude of the gradient even for large deviations to improve training stability.

\section{Experiments}
\label{sec:exp}

To assess the effectiveness of our approach, we aim to address the following questions:
\begin{itemize}
    \item How does learning in differentiable simulation compare to model-free RL in terms of sample efficiency and training time?
    \item Can flight control policies trained in differentiable simulation be deployed in the real world?
    \item How does using a simple dynamics model for BPTT affect the learning?
    \item Does pretraining state representations help improve policy learning in differentiable simulation?
\end{itemize}

The task is to recover the quadrotor from a random initial condition and stabilize it.
In the first experiment, the actor has access to the quadrotor state. In the second one, only visual features and past actions are observed (see~\ref{sec:obs}).
We compare training in differentiable simulation using BPTT with PPO~\cite{schulman2017ppo}, a widely used model-free algorithms that is well-known to work well when large-scale simulation can be exploited.
In the experiments, we include runs with different numbers of parallel environments to examine the gradient variance.
All training procedures were run on a workstation with an Nvidia Titan RTX (24 GB VRAM).

\subsection{Learning State-based Control}

\begin{figure}[t]
    \centering
    \begin{tikzpicture}
        \node at (0,0) {\includegraphics[width=1.0\linewidth]{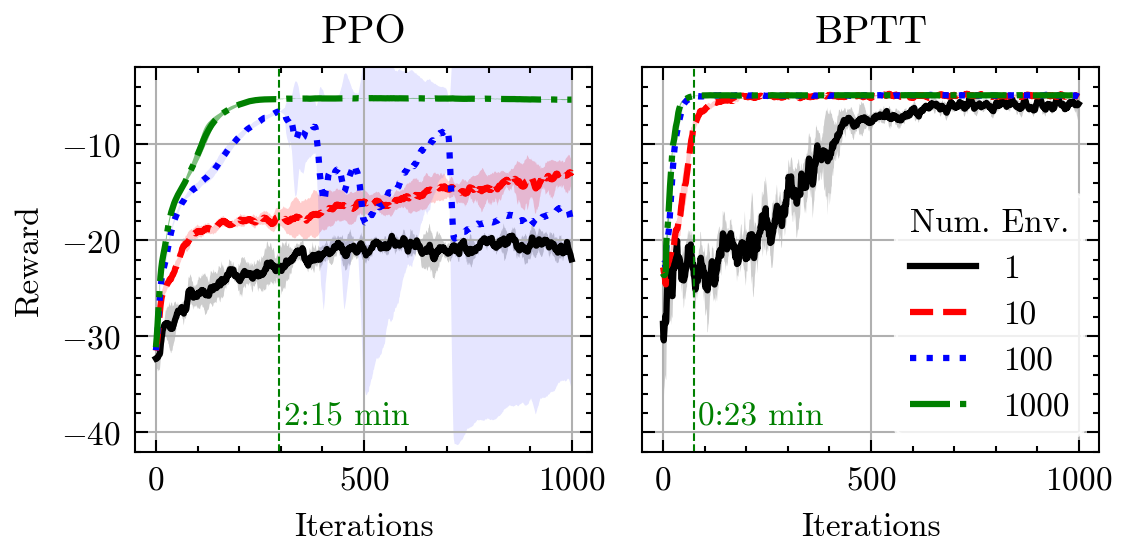}};
        
        
        \node at(0, -3.4){
        \adjustbox{max width=0.65\linewidth}{
            \begin{tabular}{c c  c | c  c}
                \toprule
                \multirow{2}{*}{Reward Target} 
                 & \multicolumn{2}{c}{Samples [Mio.]} & \multicolumn{2}{c}{Time [min]} \\  \cline{2-5} 
                 & PPO & BPTT & PPO & BPTT \\
                \midrule
                -15  &  8.4 & 2.55   & 0:25  & 0:05 \\
                -11  & 14.7 & 3.45   & 0:44 & 0:07 \\
                -7   & 22.8   & 5.4  &  1:09  & 0:11 \\
                -5.3 & 44.4   & 8.25 & 2:14   & 0:17 \\
                \bottomrule
            \end{tabular}
        }};
    \end{tikzpicture}
    \caption{Comparisons of learning state-based control using PPO and BPTT.}
    \label{fig:state-ppo-vs-bptt}
\end{figure}

We train the control policy for 1000 iterations using BPTT and PPO with 1 to 1000 parallel environments as shown in \autoref{fig:state-ppo-vs-bptt}.
We observe that using PPO, only the experiment with 1000 parallel environments stably converges.
On the other hand, all BPTT runs with more than 1 environment converge to very similar high rewards, and even with 1 single environment, the learning curve almost reaches the same performance. This result confirms the practical advantage of a low-variance gradient estimate.

Regarding the training time, using 1000 parallel environments gives the best results for both methods. Here, PPO converges after 2:15~minutes, while BPTT needs only 23~seconds.

The table in \autoref{fig:state-ppo-vs-bptt} compares how many samples and how much training time is needed to reach certain reward targets (with 1000 parallel environments).
We find that BPTT reaches all targets significantly faster and requires only a fraction of samples.
In particular, BPTT demonstrates an almost 8-times speed-up to reach the highest reward target -5.3 and requires less than 20\% of samples.

\subsection{Feature-based Control without State Estimation}

\begin{figure}[t]
    \centering
    \begin{tikzpicture}
        \node at (0,0) {\includegraphics[width=1.0\linewidth]{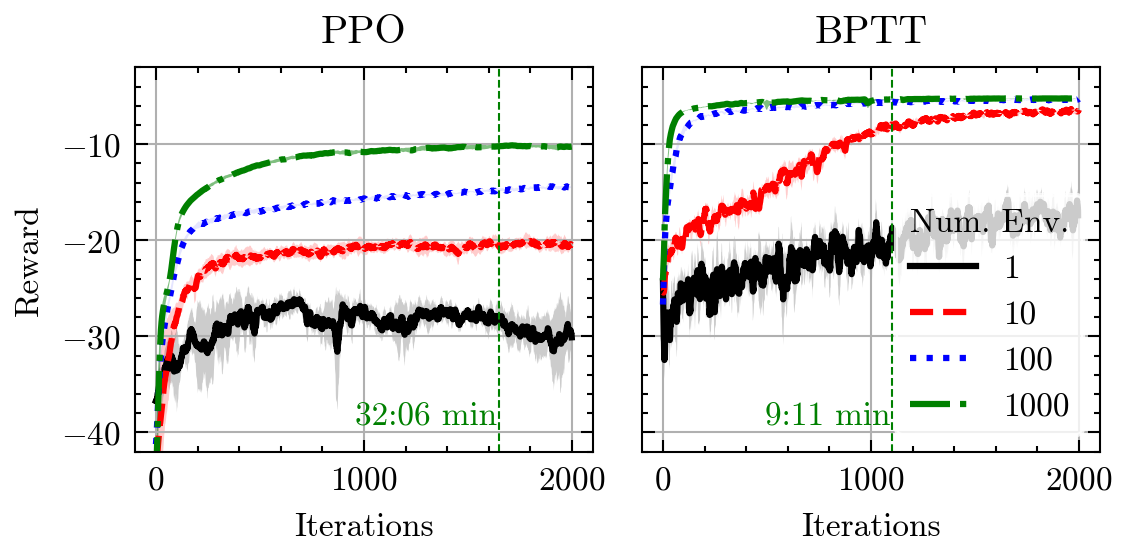}};
        
        
        \node at(0, -3.4){
        \adjustbox{max width=0.65\linewidth}{
            \begin{tabular}{c c  c | c  c}
                \toprule
                \multirow{2}{*}{Reward Target} 
                 & \multicolumn{2}{c}{Samples [Mio.]} & \multicolumn{2}{c}{Time [min]} \\  \cline{2-5} 
                 & PPO & BPTT & PPO & BPTT \\
                \midrule
                -15 &  32.1 & 2.25  & 4:10 & 0:07 \\
                -11 & 118.2 & 3.9   & 15:21 & 0:13 \\
                -7 & $-$ & 11.25 &  $-$  & 0:37 \\
                -5.3 & $-$ & 179.85   & $-$ & 10:00 \\
                \bottomrule
            \end{tabular}
        }};
    \end{tikzpicture}
    \caption{Comparisons of learning feature-based control using PPO and BPTT.}
    \label{fig:features-ppo-vs-bptt}
\end{figure}

For the more challenging task of vision-based quadrotor control, we pretrain the policy networks using state representation learning on 100,000 samples for 500 epochs (pretraining time is around 1:30 min) and subsequently run policy optimization for 2000 iterations. Unsurprisingly, the time to convergence is higher in all cases. Only BPTT with 100 or 1000 parallel environments manages to converge to a performance close to the state-based case.
Even disregarding the significantly lower reward of PPO policies at convergence, the convergence time of BPTT, 9:11~min, is more than 3 times lower compared to 32:06~min, the convergence time of PPO.
Furthermore, the table in \autoref{fig:features-ppo-vs-bptt} displays the wall-clock training time and simulation samples required to reach the reward targets. Again, we observe much lower sample and time requirements. For the last target hit by PPO, a reward of -11, BPTT was over 70 times faster and needed less than 4\% of samples.

\subsection{Sim-to-real Transfer}

\begin{figure}
    \centering
    \includegraphics[width=1\linewidth]{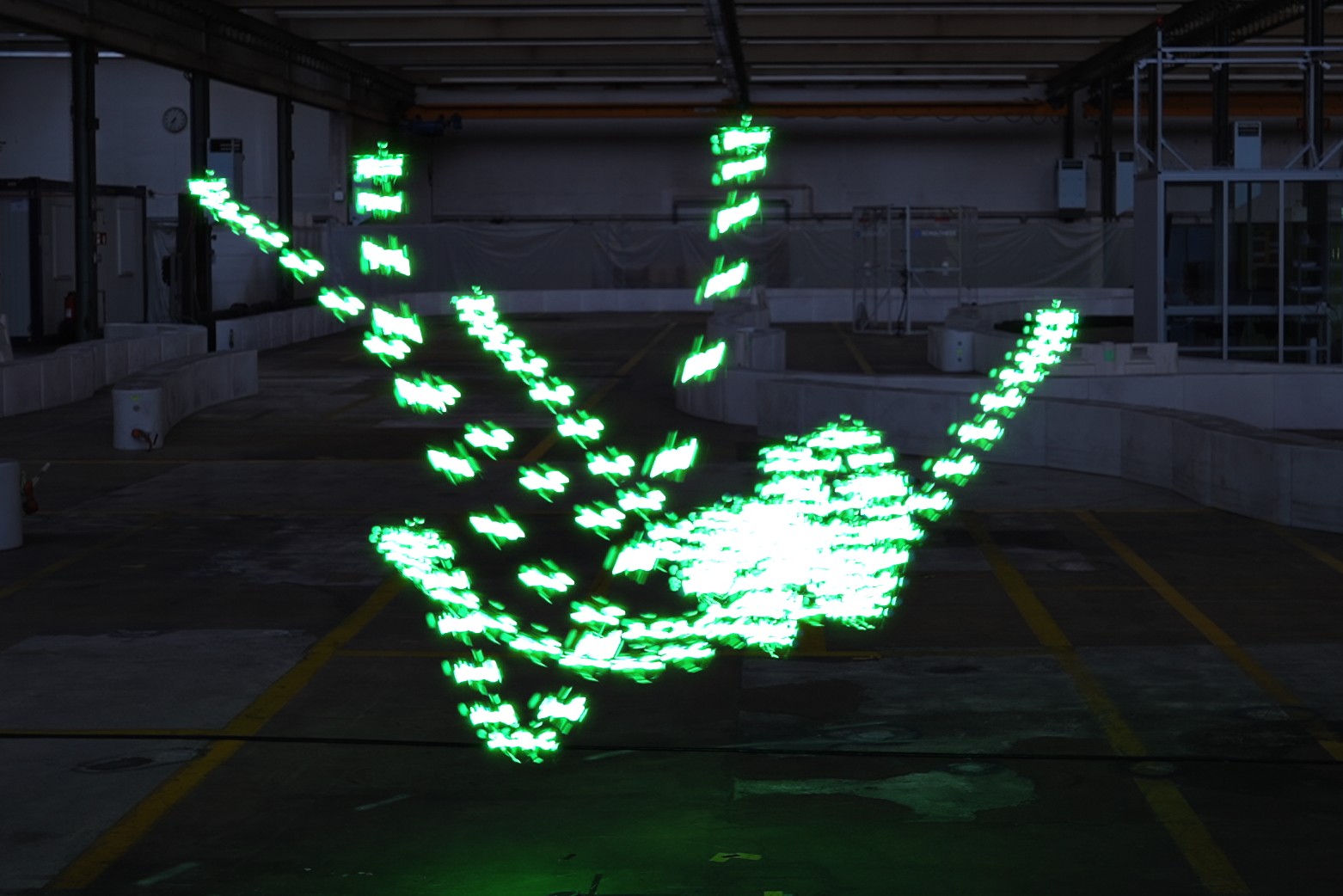}
    \caption{Feature-based control experiment (HITL). Visualization of flight trajectories with randomly initialized drone configurations.}
    \label{fig:green-traces}
\end{figure}

\begin{figure}
    \centering
    \includegraphics[width=1.0\linewidth]{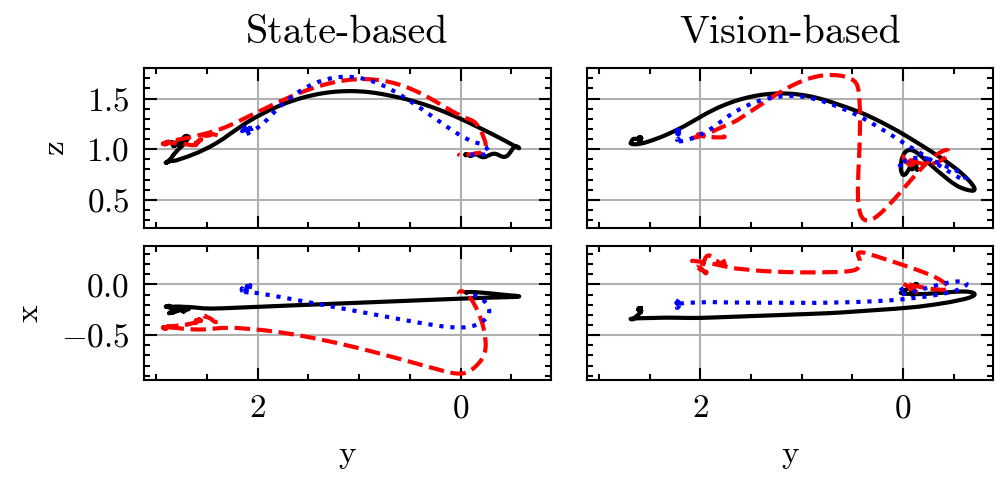}
    \caption{Real-world trajectories showing stabilization from a manual throw. Each line represents an individual trial.}
    \label{fig:real-world-trah}
\end{figure}

For real-world experiments, we use a lightweight drone built with off-the-shelf components. 
Using hardware-in-the-loop (HITL) simulation, the pose of the physical quadrotor is monitored by a motion capture system, and the actor receives virtual observations, visualized in \autoref{fig:fig1}.
The HITL experiments allow us to focus on testing the control performance without the need for a feature detection module.
Note that even though the simulated observation model causes no sim-to-real gap, the measurement noise of the motion capture systems results in out-of-distribution observations.

\autoref{fig:green-traces} illustrates the deployment of a vision-based control policy trained in differentiable simulation. The policy successfully navigates and stabilizes the quadrotor at the desired goal position.
\autoref{fig:fig1} shows a second experiment where the quadrotor is thrown manually resulting in diverse and highly unstable initial states.
As indicated by \autoref{fig:real-world-trah}, both state-based and vision-based control policies successfully perform the task in successive trials without failure.
The state-based policy stabilizes the quadrotor faster and with less oscillations around the desired position.
In particular, in the third vision-based trial (denoted in red), the quadrotor came close to the floor before approaching the goal position.
This behavior indicates that the vision-based performance could benefit from an improved observation noise model to reduce the sim-to-real gap.

\subsection{Ablation Studies}

\subsubsection{Policy Optimization Using Surrogate Gradients}

\begin{figure}
    \centering
    \includegraphics[width=1.0\linewidth]{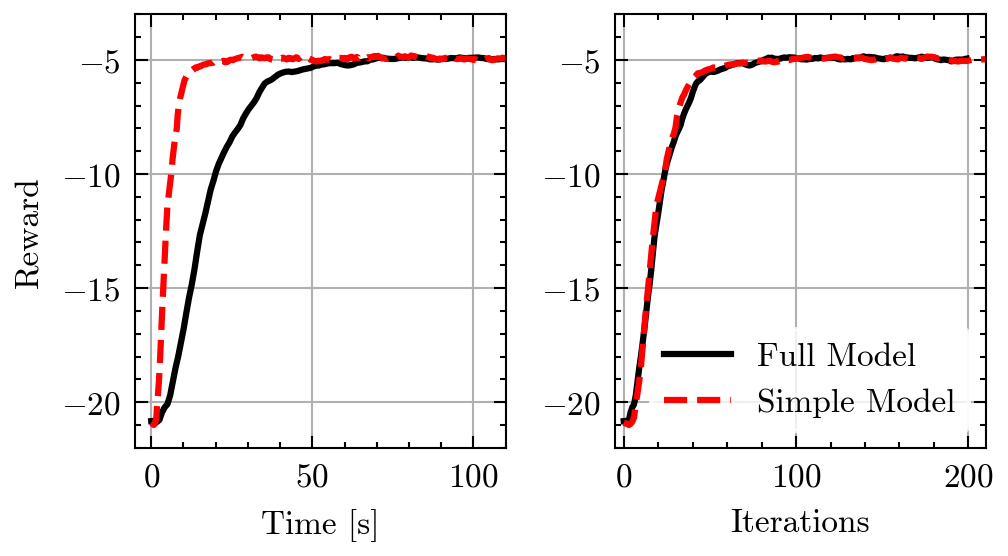}
    \caption{Using a simple model for gradient computations reduces training time without sacrificing on performance.}
    \label{fig:ablation-model}
\end{figure}

Using a simpler dynamics model for the backward path directly increases the number of simulation steps per second, as shown in \autoref{tab:sim-speed}.
However, using only a surrogate gradient might weaken the quality of the gradient and hence, decrease sample efficiency.
\autoref{fig:ablation-model} shows two runs with BPTT over 200 iterations (100 environments) on the state-based control task. Both runs use the same initial parameters and random seeds. The only difference is the quadrotor model.
It clearly shows that the simple model speeds up training but does not sacrifice sample efficiency or performance.

\subsubsection{State Representation Learning as Pretraining}

\begin{figure}
    \centering
    \includegraphics[width=0.8\linewidth]{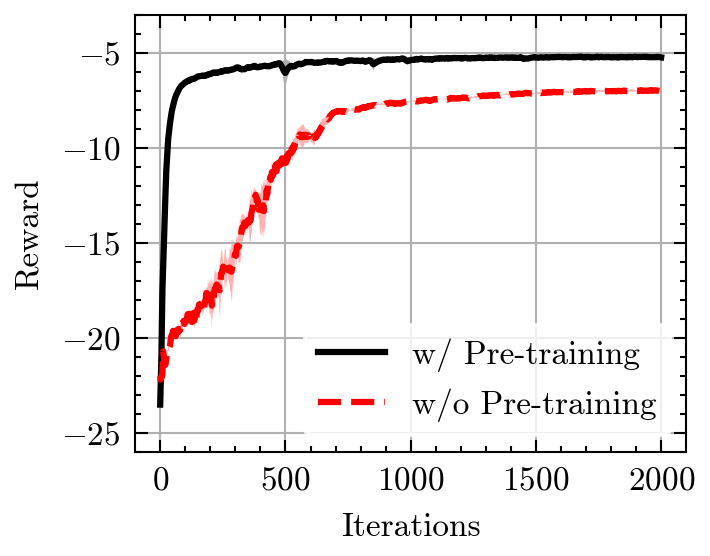}
    \caption{Pre-training the neural network parameters on state representation learning improves convergence and performance during policy optimization.}
    \label{fig:ablation-pretraining}
\end{figure}

\autoref{fig:ablation-pretraining} compares the BPTT training curves on the vision-based control task with and without state representation pretraining using 1000 environments.
We observe that pretraining improves both convergence and asymptotic performance significantly. Note, however, that even without pretraining, the performance surpasses the one of PPO on the same task.

We hypothesize that state representation pretraining is particularly useful when combined with differentiable simulation because it not only initializes the weights to provide a more meaningful internal representation but also improves the quality of the gradient.
Possibly, the pretrained weights make the first-order approximation of the policy optimization objective more accurate in a larger vicinity of the current parameters.
\section{Conclusion}

This paper investigated using differentiable simulation to learn quadrotor control policies from state and visual observations.
We found that differentiable simulation can accelerate policy training drastically, leading to more than 7 times faster training speeds and requiring less than 20\% of samples compared to model-free reinforcement learning.
We show that by using a simple dynamics model, computational costs can be significantly reduced without compromising on sample efficiency.
Moreover, leveraging pretraining on a state representation task helps convergence and improves performance substantially.

Our work demonstrates the first application of differentiable simulation to low-level quadrotor control. We deploy the trained policies successfully in the real world on the challenging task of stabilizing the quadrotor after a manual throw without state estimation and only observing visual features.

Overall, the results suggest that differentiable simulation has the potential of introducing a paradigm shift for learning in simulation.
We believe that despite the increasing efforts on developing high-fidelity differentiable simulators, using simple backward models will be faster and sufficient in many cases and requires only limited engineering effort.
Therefore, we hope to inspire the community to try differentiable simulation and accelerate their research and innovation.



\newpage

\balance

\bibliographystyle{IEEEtran}
\typeout{} 
\bibliography{references}

@STRING{ieee    = "Proc. {IEEE}" }

@STRING{iros    = "IEEE/RSJ Int. Conf. Intell. Robot. Syst. (IROS)" }

@STRING{arxiv   = "ar{X}iv e-prints" }

@inproceedings{kolter2009,
  title = {Policy Search via the Signed Derivative},
  booktitle = {Robotics: {{Science}} and {{Systems V}}},
  author = {Kolter, J. Z. and Ng, A. Y.},
  year = {2009},
  month = jun,
  publisher = {{Robotics: Science and Systems Foundation}},
  doi = {10.15607/RSS.2009.V.027},
  urldate = {2025-02-28},
  isbn = {978-0-262-51463-7},
}

@inproceedings{abbeel2006,
  title = {Using Inaccurate Models in Reinforcement Learning},
  booktitle = {Proceedings of the 23rd International Conference on {{Machine}} Learning  - {{ICML}} '06},
  author = {Abbeel, Pieter and Quigley, Morgan and Ng, Andrew Y.},
  year = {2006},
  pages = {1--8},
  publisher = {ACM Press},
  address = {Pittsburgh, Pennsylvania},
  doi = {10.1145/1143844.1143845},
  urldate = {2025-02-28},
  isbn = {978-1-59593-383-6},
  langid = {english},
}

@article{haarnoja2024,
  title = {Learning Agile Soccer Skills for a Bipedal Robot with Deep Reinforcement Learning},
  author = {Haarnoja, Tuomas and Moran, Ben and Lever, Guy and Huang, Sandy H. and Tirumala, Dhruva and Humplik, Jan and Wulfmeier, Markus and Tunyasuvunakool, Saran and Siegel, Noah Y. and Hafner, Roland and Bloesch, Michael and Hartikainen, Kristian and Byravan, Arunkumar and Hasenclever, Leonard and Tassa, Yuval and Sadeghi, Fereshteh and Batchelor, Nathan and Casarini, Federico and Saliceti, Stefano and Game, Charles and Sreendra, Neil and Patel, Kushal and Gwira, Marlon and Huber, Andrea and Hurley, Nicole and Nori, Francesco and Hadsell, Raia and Heess, Nicolas},
  year = {2024},
  month = apr,
  journal = {Science Robotics},
  volume = {9},
  number = {89},
  pages = {eadi8022},
  issn = {2470-9476},
  doi = {10.1126/scirobotics.adi8022},
  urldate = {2024-05-07},
  langid = {english},
}

@article{metz2021gradients,
  title={Gradients are not all you need},
  author={Metz, Luke and Freeman, C Daniel and Schoenholz, Samuel S and Kachman, Tal},
  journal={arXiv preprint arXiv:2111.05803},
  year={2021}
}

@inproceedings{xu2021accelerated,
    title={Accelerated Policy Learning with Parallel Differentiable Simulation},
    author={Xu, Jie and Makoviychuk, Viktor and Narang, Yashraj and Ramos, Fabio and Matusik, Wojciech and Garg, Animesh and Macklin, Miles},
    booktitle={International Conference on Learning Representations},
    year={2021}
}

@article{hwangbo2019learning,
  title={Learning agile and dynamic motor skills for legged robots},
  author={Hwangbo, Jemin and Lee, Joonho and Dosovitskiy, Alexey and Bellicoso, Dario and Tsounis, Vassilios and Koltun, Vladlen and Hutter, Marco},
  journal={Science Robotics},
  volume={4},
  number={26},
  pages={eaau5872},
  year={2019},
  publisher={American Association for the Advancement of Science}
}

@article{freeman2021brax,
  title={Brax--A Differentiable Physics Engine for Large Scale Rigid Body Simulation},
  author={Freeman, C Daniel and Frey, Erik and Raichuk, Anton and Girgin, Sertan and Mordatch, Igor and Bachem, Olivier},
  journal={arXiv preprint arXiv:2106.13281},
  year={2021}
}

@misc{jax2018github,
  author = {James Bradbury and Roy Frostig and Peter Hawkins and Matthew James Johnson and Chris Leary and Dougal Maclaurin and George Necula and Adam Paszke and Jake Vander{P}las and Skye Wanderman-{M}ilne and Qiao Zhang},
  title = {{JAX}: composable transformations of {P}ython+{N}um{P}y programs},
  url = {http://github.com/google/jax},
  version = {0.3.13},
  year = {2018},
}

@article{howelllecleach2022,
  title={Dojo: A Differentiable Simulator for Robotics},
  author={Howell, Taylor and Le Cleac'h, Simon and Bruedigam, Jan and Kolter, Zico and Schwager, Mac and Manchester, Zachary},
  journal={arXiv preprint arXiv:2203.00806},
  year={2022}
}

@inproceedings{song2020flightmare,
  title={Flightmare: A Flexible Quadrotor Simulator},
  author={Song, Yunlong and Naji, Selim and Kaufmann, Elia and Loquercio, Antonio and Scaramuzza, Davide},
  booktitle = {Conference on Robot Learning {(CoRL)}},
  year = {2020},
}

@article{hwangbo2017control,
  title={Control of a quadrotor with reinforcement learning},
  author={Hwangbo, Jemin and Sa, Inkyu and Siegwart, Roland and Hutter, Marco},
  journal={IEEE Robotics and Automation Letters},
  volume={2},
  number={4},
  pages={2096--2103},
  year={2017},
  publisher={IEEE}
}

@inproceedings{song2021autonomousdrone,
	title={Autonomous Drone Racing with Deep Reinforcement Learning},
	author={Song, Yunlong and Steinweg, Mats and Kaufmann, Elia and Scaramuzza, Davide},
	booktitle={2021 IEEE/RSJ International Conference on Intelligent Robots and Systems (IROS)},
	pages={1205--1212},
	year={2021},
	organization={IEEE}
}

@article{song2023reaching,
	title={Reaching the limit in autonomous racing: Optimal control versus reinforcement learning},
	author={Song, Yunlong and Romero, Angel and M{\"u}ller, Matthias and Koltun, Vladlen and Scaramuzza, Davide},
	journal={Science Robotics},
	volume={8},
	number={82},
	pages={eadg1462},
	year={2023},
	publisher={American Association for the Advancement of Science}
}

@article{song2024learning,
  title={Learning Quadruped Locomotion Using Differentiable Simulation},
  author={Song, Yunlong and Kim, Sangbae and Scaramuzza, Davide},
  journal={arXiv preprint arXiv:2403.14864},
  year={2024}
}

@article{kaufmann2023champion,
  title={Champion-level drone racing using deep reinforcement learning},
  author={Kaufmann, Elia and Bauersfeld, Leonard and Loquercio, Antonio and M{\"u}ller, Matthias and Koltun, Vladlen and Scaramuzza, Davide},
  journal={Nature},
  volume={620},
  number={7976},
  pages={982--987},
  year={2023},
  publisher={Nature Publishing Group UK London}
}

@article{geles2024demonstrating,
  title={Demonstrating Agile Flight from Pixels without State Estimation},
  author={Geles, Ismail and Bauersfeld, Leonard and Romero, Angel and Xing, Jiaxu and Scaramuzza, Davide},
  journal={RSS: Robotics, Science, and Systems},
  year={2024},
  publisher={IEEE}
}

@inproceedings{usenko2018doublesqhere,
	title = {The Double Sphere Camera Model},
	url = {http://arxiv.org/abs/1807.08957},
	doi = {10.1109/3DV.2018.00069},
	urldate = {2024-08-26},
	booktitle = {2018 {International} {Conference} on {3D} {Vision} ({3DV})},
	author = {Usenko, Vladyslav and Demmel, Nikolaus and Cremers, Daniel},
	month = sep,
	year = {2018},
	note = {arXiv:1807.08957 [cs]},
	pages = {552--560},
}

@misc{schulman2017ppo,
	title = {Proximal {Policy} {Optimization} {Algorithms}},
	url = {http://arxiv.org/abs/1707.06347},
	urldate = {2023-01-20},
	author = {Schulman, John and Wolski, Filip and Dhariwal, Prafulla and Radford, Alec and Klimov, Oleg},
	month = aug,
	year = {2017},
	note = {9994 citations (Semantic Scholar/arXiv) [2023-11-26]
arXiv:1707.06347 [cs]},
}

@article{debruin2018representation,
	title = {Integrating {State} {Representation} {Learning} {Into} {Deep} {Reinforcement} {Learning}},
	volume = {3},
	issn = {2377-3766},
	url = {https://ieeexplore.ieee.org/abstract/document/8276247},
	doi = {10.1109/LRA.2018.2800101},
	number = {3},
	urldate = {2024-09-15},
	journal = {IEEE Robotics and Automation Letters},
	author = {de Bruin, Tim and Kober, Jens and Tuyls, Karl and Babuška, Robert},
	month = jul,
	year = {2018},
	note = {Conference Name: IEEE Robotics and Automation Letters},
	pages = {1394--1401},
}

@article{eschmann2024learning,
  title={Learning to fly in seconds},
  author={Eschmann, Jonas and Albani, Dario and Loianno, Giuseppe},
  journal={IEEE Robotics and Automation Letters},
  year={2024},
  publisher={IEEE}
}

@article{zhang2024back,
  title={Back to Newton's Laws: Learning Vision-based Agile Flight via Differentiable Physics},
  author={Zhang, Yuang and Hu, Yu and Song, Yunlong and Zou, Danping and Lin, Weiyao},
  journal={arXiv preprint arXiv:2407.10648},
  year={2024}
}

@misc{tang2024rlrobotics,
	title = {Deep {Reinforcement} {Learning} for {Robotics}: {A} {Survey} of {Real}-{World} {Successes}},
	shorttitle = {Deep {Reinforcement} {Learning} for {Robotics}},
	url = {http://arxiv.org/abs/2408.03539},
	doi = {10.48550/arXiv.2408.03539},
	urldate = {2024-08-30},
	publisher = {arXiv},
	author = {Tang, Chen and Abbatematteo, Ben and Hu, Jiaheng and Chandra, Rohan and Martín-Martín, Roberto and Stone, Peter},
	month = aug,
	year = {2024},
	note = {arXiv:2408.03539 [cs]},
}

%
\end{document}